\documentclass[letterpaper]{article} 
\usepackage{aaai25}  
\usepackage{times}  
\usepackage{helvet}  
\usepackage{courier}  
\usepackage[hyphens]{url}  
\usepackage{graphicx} 
\usepackage{natbib}  
\usepackage{caption} 
\usepackage{algorithm}
\usepackage{algorithmic}
\usepackage{enumitem}

\usepackage{newfloat}
\usepackage{listings}
\DeclareCaptionStyle{ruled}{labelfont=normalfont,labelsep=colon,strut=off} 

\floatstyle{ruled}
\newfloat{listing}{tb}{lst}{}
\floatname{listing}{Listing}
\usepackage{booktabs}
\usepackage{amsmath,amssymb,amsfonts}
\DeclareMathOperator*{\argmax}{arg\,max}

\def\BibTeX{{\rm B\kern-.05em{\sc i\kern-.025em b}\kern-.08em
    T\kern-.1667em\lower.7ex\hbox{E}\kern-.125emX}}

\title{LLM-based Corroborating and Refuting Evidence Retrieval \\for Scientific Claim Verification}
\author{
    Siyuan Wang\textsuperscript{\rm 1}, 
    James R. Foulds\textsuperscript{\rm 2}, 
    Md Osman Gani\textsuperscript{\rm 2}, 
    Shimei Pan\textsuperscript{\rm 2}\thanks{Corresponding author.}
}
\affiliations{
    \textsuperscript{\rm 1}Anhui Normal University, China\\
    \textsuperscript{\rm 2}University of Maryland, Baltimore County, USA\\
    wangtif@ahnu.edu.cn, jfoulds@umbc.edu, mogani@umbc.edu, shimei@umbc.edu
}

\begin{document}

\maketitle

\begin{abstract}
    In this paper, we introduce CIBER (Claim Investigation Based on Evidence Retrieval), an extension of the Retrieval-Augmented Generation (RAG) framework designed to identify corroborating and refuting documents as evidence for scientific claim verification. CIBER addresses the inherent uncertainty in Large Language Models (LLMs) by evaluating response consistency across diverse interrogation probes. By focusing on the behavioral analysis of LLMs without requiring access to their internal information, CIBER is applicable to both white-box and black-box models. Furthermore, CIBER operates in an unsupervised manner, enabling easy generalization across various scientific domains. Comprehensive evaluations conducted using LLMs with varying levels of domain expertise and linguistic proficiency reveal CIBER's superior performance compared to conventional RAG approaches. These findings not only highlight the effectiveness of CIBER but also provide valuable insights for future advancements in LLM-based scientific claim verification.
\end{abstract}

\section{Introduction}

\begin{figure*}[ht]
\centering
\includegraphics[width=0.74\textwidth]{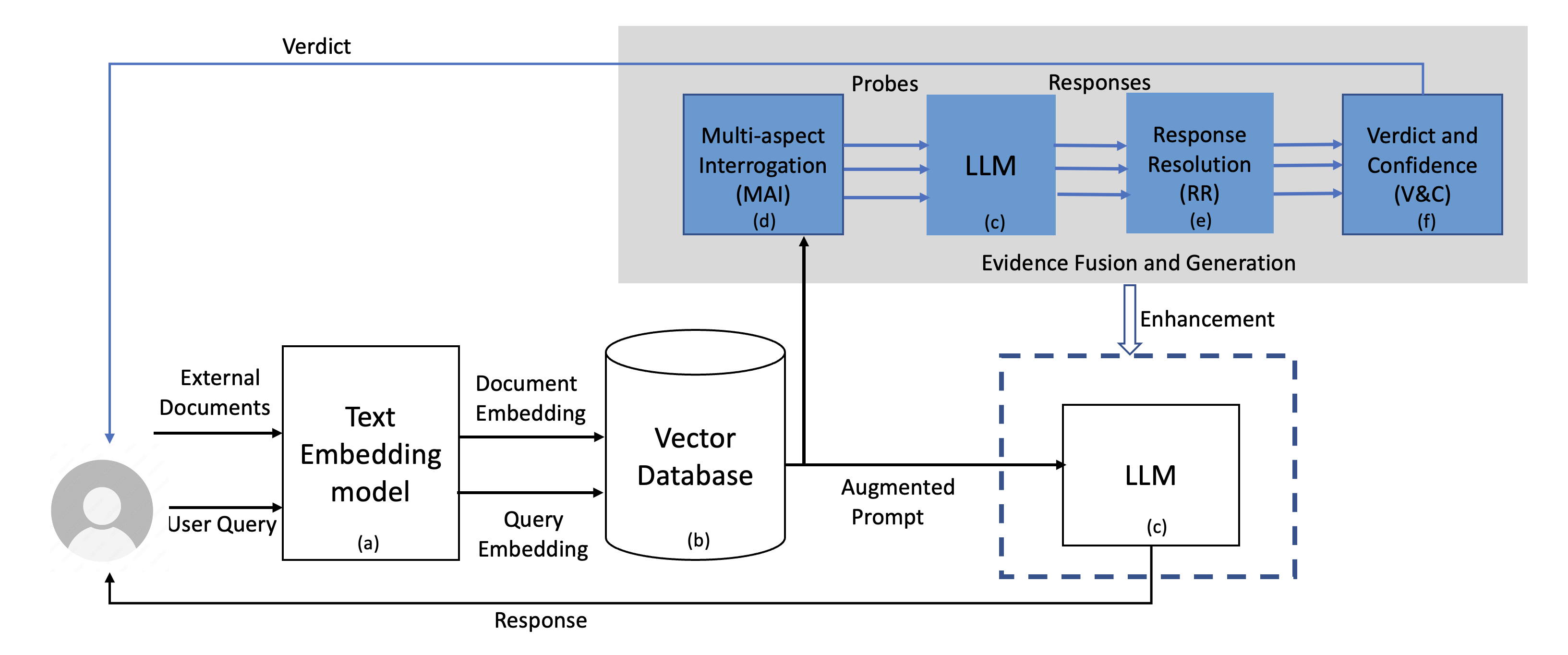}
\caption{CIBER Architecture}
\label{fig:BRAG}
\end{figure*}

Recent advances in large language model (LLM) technology promise to redefine how we interact with language technologies and use them in the new digital era~\cite{Panagoulias2023,Wei2023,Guo2023,Yang2023}. One common challenge faced by LLMs is ``hallucination'' where they may produce answers that seem correct but are actually inaccurate or misleading~\cite{zhang2023sirens}. This can be particularly problematic in scientific investigations where accuracy and reliability of evidences and claims are critical. Hallucinations often arise when the necessary knowledge to answer a specific user query is absent or inadequate from an LLM's training data. Retrieval-Augmented Generation (RAG), a technique designed to improve the reliability of LLMs~\cite{lewis2020retrieval}, can mitigate this issue by dynamically bringing in external knowledge sources relevant to a user's query. Prior research has demonstrated that RAG can significantly reduce the hallucination rate of LLMs and enhance their response reliability~\cite{feldman2023trapping,feldman2024ragged}. Since its debut, RAG has rapidly gained popularity~\cite{siriwardhana2023improving,chen2024benchmarking,hofstatter2023fid} and been integrated into numerous AI applications and services. Although RAG can significantly reduce hallucination, the accuracy of LLM responses can still fluctuate due to factors such as the topic and wording of user queries, the relevance and consistency of retrieved documents, and the language comprehension and reasoning abilities of the LLM model employed, potentially resulting in errors in responses. For instance, extensive evaluation on benchmark datasets indicates that RAG still struggles regarding noise robustness, negative rejection, and information integration ~\cite{chen2024benchmarking}. 

To further reduce the errors in RAG, we introduce CIBER, designed to identify and retrieve scientific documents as corroborating or refuting evidences for claim verification. A claim could range from causal statements like ``Human activities may  cause climate change'' to hypotheses such as ``Dysregulation of microRNA expression may contribute  to the pathogenesis of cardiovascular diseases such as hypertension.'' Given a claim $C$ by a user (e.g., a scientist), CIBER can automatically (1) retrieve relevant scientific publications from reliable sources (e.g., The New England Journal of Medicine for biomedical research or NeurIPS or ICML  for Machine Learning), (2) validate the claim $C$ against each of the retrieved publications (e.g. full papers or abstracts) through multi-faceted interrogations, providing a verdict  on its truthfulness along with a confidence score, and (3) present the most representative papers that either support or refute the claim. 

Figure~\ref{fig:BRAG} depicts the system diagram of CIBER, which is an extension of a typical RAG architecture that includes (1) a text embedding model (Figure~\ref{fig:BRAG}(a)) that converts text from both external sources (e.g., scientific publications) and user queries (e.g., a claim to be verified) into embedding vectors that capture their semantics, (2) a vector database (Figure~\ref{fig:BRAG}(b)) enabling similarity-based retrieval of relevant documents given a user query, and (3) an LLM (Figure~\ref{fig:BRAG}(c)) that analyzes the user query as well as the retrieved documents to generate an answer. In CIBER, we focus on enhancing the last stage of RAG (the new CIBER components are shown in the gray box in Figure~\ref{fig:BRAG}). Specifically, we employ multi-aspect interrogation (MAI) (Figure~\ref{fig:BRAG}(d)) to generate different probes to assess the reliability of LLM responses.  The Response Resolution (RR) module (Figure~\ref{fig:BRAG}(e)) parses the LLM responses and maps them to one of the three canonical answers: Support, Refute, and Neutral. The Verdict and Confidence (V\&C) module (Figure~\ref{fig:BRAG}(f)) aggregates all the responses from all the probes and determines whether the input claim $C$ is supported or refuted by combined evidence with a confidence score. Documents with the highest confidence scores can be presented to the user as representative work that either supports or refutes the claim. The main contributions of our project include: 

\begin{itemize}
    \item Propose a new CIBER framework that can further reduce hallucination in LLM generation than a typical RAG. CIBER is unsupervised, making it applicable in diverse scientific fields. Moreover, it does not require access to LLM internal information (e.g., model parameters or training data), making it suitable for both white-box and black-box LLMs. 
    \item Develop various methods to systematically integrate and fuse evidences gathered from different interrogation probes to determine the truthfulness of a claim, along with an associated confidence score.
    \item Create two new synthetic and two new real datasets to assess the effectiveness of the proposed method.
\end{itemize}

\section{Related Work}
RAG has experienced significant development in recent years. Initially, RAG systems focused on directly augmenting Large Language Models (LLMs) with retrieved knowledge through enhanced pre-training techniques~\cite{lewis2020retrieval,borgeaud2022improving}. However, with advanced LLMs demonstrating strong contextual learning abilities, RAG research has transitioned towards providing improved and more relevant contextual information. These retrieval-based enhancements include improved source selection~\cite{li2023classification} and query expansion~\cite{ma2023query,peng2023large}, refined content indexing~\cite{wang2024knowledge}, enhanced content ranking~\cite{zhuang2023open}, and advanced iterative and recursive retrieval techniques~\cite{shao2023enhancing,trivedi2022interleaving}. In contrast, our focus is on enhancing the generation stage of RAG where we systematically measure the uncertainty in LLM responses to a variety of interrogation probes to determine the reliability of these responses and determine the final verdicts. 

There is also a large body of work on non-RAG-based automated claim verification. It often involves four stages, beginning with claim detection, where claims are selected for verification based on their check-worthiness~\cite{Hassan2015}.  Evidence retrieval aims to find evidences to indicate a claim's veracity, often using metadata or stance detection techniques ~\cite{Ferreira2016,Hanselowski2019}. Verdict prediction determines the truthfulness of claims~\cite{Nakashole2014,Augenstein2019}. Additionally, knowledge graph-based fact verification have been proposed to assess the veracity of extracted claims by leveraging structured knowledge bases~\cite{tchechmedjiev2019claimskg}. Finally, justification generation explains the reasoning behind verdicts, with strategies including attention weights, logic-based systems, or textual explanations~\cite{Popat2018,Ahmadi2019,Atanasova2024}.  Among the four tasks in the claim extraction and verification pipeline, we only focus on  evidence retrieval and verdict prediction. 

\section{Research Questions}
In this study, we focus on three research questions:
\begin{itemize}[leftmargin=*]
\item (RQ1) How does the performance of CIBER compare with that of a typical RAG? How does its performance vary with different LLMs with diverse language understanding and reasoning abilities?
\item (RQ2) How do various interrogation strategies within the MAI module influence the performance of CIBER?
\item (RQ3) What effects do different evidence fusion strategies within the V\&C module have on CIBER performance?

\end{itemize}

In the following, we provide details on the design and implementation of CIBER.

\begin{table*}[th]
\centering
\begin{small}
    \begin{tabular}{cccc}
        \toprule
            &\multicolumn{3}{c}{Verdict} \\
        \cmidrule{2-4}
            & Support & Refute & Neutral\\
        \midrule
        $P_{AG}$ & $Prob(r_{i}=S)$ & $Prob(r_{i}=R)$ & $Prob(r_{i}=N)$ \\
        $P_{CF}$ & $\alpha*Prob(r_{i}=S)$ & $\alpha*Prob(r_{i}=R)$ & $Prob(r_{i}=N)$ + $(1-\alpha)$*$\big (Prob(r_{i}=S)$ + $Prob(r_{i}=R)$\big )\\
        \bottomrule
    \end{tabular}
\end{small}
 \caption{Mass function $m(\cdot)$ used in Dempster-Shafer Belief Update.}
\label{mass_func}
\end{table*}

\section{Methodology}
In this section, we explain the main CIBER modules including  MAI,  RR, and a V\&C.

\subsection{Multi-Aspect Interrogation (MAI)}
MAI is designed to assess the consistency of LLM responses under various probes with diverse lexical and logical variations. MAI begins by verifying a claim $C$ within the context of each retrieved paper/abstract. This step is critical in verifying scientific claims, considering the specialized knowledge necessary to either understand or verify such claims may not be adequately represented in conventional LLM training datasets. By anchoring the LLM's responses in a specific scientific study, this contextual grounding enables the LLM to provide more accurate and relevant information in response to specific scientific claims. Specifically, given an input claim $C$, a retrieved publication $A$  and an LLM $L$, we construct the Original Probe $p_{O}$ (e.g., ``Based on the study presented in Paper A, is Claim C true?'') We also create additional probes to interrogate $L$ from various perspectives. To test $L$'s logic consistency, we design two sets of probes: $P_{AG}$ (``AGree Probe'') and $P_{CF}$ (``ConFlict Probe.'') $P_{AG}$ includes probes whose LLM responses should align (or agree) with those from $p_{O}$, while $P_{CF}$ contains probes whose responses should contradict (or disagree with) those from $p_{O}$.  Moreover, to test $L$'s ability in understanding user queries with different lexical and syntactic variations, for each $p_{i}$ in either $P_{AG}$ or $P_{CF}$, we add $g(p_{i})$, which is a paraphrase of $p_{i}$. To illustrate, imagine a climate researcher seeking to understand the causal link between human activities and climate change. Instead of directly querying an LLM like ChatGPT ``Can human activities cause climate change?'', which may yield inaccurate results, our system begins by accessing papers from high-quality venues (e.g., the journal of Nature Climate Change).  For each paper/abstract, the system examines the LLM responses from various perspectives.  In this example, $p_{O}$ can be a query such as ``Based on the study described in paper $A$, is the claim $C$ true?'' We can populate $P_{AG}$ with $p_{O}$ plus paraphrases of {$p_{O}$} such as ``Is claim $C$ supported by the study described in paper $A$.''  We can populate $P_{CF}$ with $\neg p_{O}$ such as ``Based on the study described in paper $A$, is the claim $C$ false?'' and paraphrases of $\neg p_{O}$ such as ``Is claim $C$ refuted by the study described in paper $A$.'' 

\subsection{Response Resolution (RR)}

We employ specific prompt engineering strategies to facilitate the parsing of LLM responses. For GPT-2, which operates in a sentence completion mode, we append adverbs like ``relatively'' or ``quite'' in the prompts so that the words completed by the LLM are more restricted. For instance, in the prompt ``Based on the study described in the paper, the likelihood that 'human activities may cause climate change' is \textbf{relatively/quite} [Blank],'' the appended adverbs significantly limit the potential LLM responses to words like ``high'' or ``low.'' In contrast, without these adverbs, the expressions in [Blank] could vary widely, ranging from ``contingent on extensive scientific evidence'' to ``subject to further investigation.'' Similarly, for GPT-3.5 and GPT-4, which operate in a Q\&A model, we append specific instructions to constrain the response format: ``Please answer with either yes or no. If you are not sure, please say 'I am not sure'.'' While these prompt engineering strategies work most of the time, there's no guarantee that the LLMs will always follow the instructions. For example,  occasionally GPT-3.5/GPT-4' may generate a response like ``According to the abstract provided, the statement 'human activities may cause climate change' is false.''  To process these responses, we developed a straightforward  lexicon and regular expression-based parser to map these responses to their canonical forms. For example, for GPT-2, we extract the words after the adverb ``relatively'' or ``quite'' and map them to either ``Support,'' ``Refute'' or `` Neutral.''  To parse the responses from GPT-3.5 and GPT-4, we developed a regular expression-based parser to identify direct answers such as ``Yes,'' ``No'' and `` I am not sure'' or other variants such as ``is correct'' and ``is not false.''

\subsection{Verdict and Confidence (V\&C)}
Given a claim $C$ and a retrieved publication $A$, the V\&C module gathers all the LLM responses from all the probes in $P_{AG}$ and $P_{CF}$ and generates the final verdict $V$ plus a confidence score $CS$.  Given that an LLM typically employs a stochastic text generation process, sending the same probe to the LLM multiple times can result in varied responses. To address this uncertainty, we iterate each probe in $P_{AG}$ and $P_{CF}$ $K$ times, recording all the $K$ LLM responses per probe.  We explored three fusion strategies to combine the evidences from all the probes. 

\subsubsection{Weighted Proportions (WP)}
Before consolidating the evidences from all probes, we invert the LLM responses generated from $p_{i}\in P_{CF}$, because a ``Support'' response to $p_{i}\in P_{CF}$ is equivalent to a ``Refute'' response to $p_{j}\in P_{AG}$. Subsequently, we employ the following formulas to calculate the verdict $V_{WP}$.
\begin{small}
\begin{equation}   
\begin{aligned}
    WP_S= & \alpha * Prob(r_{i}=S, i \in P_{AG}) \\
          & + (1-\alpha)* Prob(r_{i}=S, i \in P_{CF}) \\
\end{aligned}
\end{equation}
\end{small}

\begin{small}
\begin{equation}   
\begin{aligned}
    WP_R= & \alpha * Prob(r_{i}=R, i \in P_{AG}) \\
          & + (1-\alpha)* Prob(r_{i}=R, i \in P_{CF}) \\
\end{aligned}
\end{equation}
\end{small}

\begin{small}
\begin{equation}   
\begin{aligned}
    WP_N= & \alpha * Prob(r_{i}=N, i \in P_{AG}) \\
          & + (1-\alpha)* Prob(r_{i}=N, i \in P_{CF}) \\
\end{aligned}
\end{equation}
\end{small}

\begin{small}
\begin{equation}        
    V_{WP}= \argmax_ {Q\in (S,R,N)}{WP_{Q}}
\end{equation}
\end{small}

where $r_{i}$ represents an LLM response generated with probe $i$, and $Prob(r_{i})$ denotes the probability of observing response $r_{i}$ among the responses. Here, $r_{i}$ can take one of three values: ``(S)upport,'' ``(R)efute,'' or ``(N)eutral''. $\alpha$ serves as a trade-off parameter that regulates the relative importance assigned to the evidence derived from the probes in $P_{AG}$ compared to those in $P_{CF}$.  The final confidence score, $CS_{WP}=WP_{V_{WP}}$.

\subsubsection{Weighted Information Gain (WIG)}
Similar to   WP, we begin by reversing the LLM responses generated from $p_{i}\in P_{CF}$. Subsequently, we utilize information theory-based metrics to evaluate the uncertainty in these LLM responses. Information theory is a mathematical framework for quantifying the amount of information in data~\cite{shannon1948mathematical}. It explores concepts such as entropy to measure the uncertainty in a dataset.  In addition to entropy $E$, we also compute Information Gain (IG) to quantify the  reduction of uncertainty given the evidences from LLM responses.  

\begin{small}
\begin{equation}
\begin{aligned}
    E_{AG} = - \sum_{r_{i}\in\{S, R, N\}}Prob(r_{i}, i\in P_{AG})\log Prob(r_{i},i\in P_{AG}) 
\end{aligned}    
\end{equation}
\end{small}

\begin{small}
\begin{equation}
\begin{aligned}
    E_{CF} =  - \sum_{r_{i}\in\{S, R, N\}}Prob(r_{i}, i\in P_{CF})\log Prob(r_{i},i\in P_{CF})
\end{aligned}    
\end{equation}
\end{small}

\begin{small}
\begin{equation}
\begin{aligned}
    IG_{AG} =  logM-E_{AG} 
\end{aligned}
\end{equation}
\end{small}

\begin{small}
\begin{equation}
\begin{aligned}
    IG_{CF} =  logM-E_{CF}   \mbox{ ,}
 \end{aligned}
\end{equation}
\end{small}

where $M$ represents the  number of possible verdicts which is 3: ``Support,'' ``Refute,'' and ``Neutral.''   Next we compute the Weighted Information Gain (WIG) and  $V_{WIG}$:

\begin{small}
\begin{equation}   
\begin{aligned}
    WIG_S= & \alpha * IG_{AG}* Prob(r_{i}=S, i \in P_{AG}) \\
          & + (1-\alpha)*IG_{CF} Prob(r_{i}=S, i \in P_{CF}) \\
\end{aligned}
\end{equation}
\end{small}

\begin{small}
\begin{equation}   
\begin{aligned}
    WIG_R= & \alpha * IG_{AG}* Prob(r_{i}=R, i \in P_{AG}) \\
          & + (1-\alpha)*IG_{CF} Prob(r_{i}=R, i \in P_{CF}) \\
\end{aligned}
\end{equation}
\end{small}

\begin{small}
\begin{equation}   
\begin{aligned}
    WIG_N= & \alpha * IG_{AG}* Prob(r_{i}=N, i \in P_{AG}) \\
          & + (1-\alpha)*IG_{CF} Prob(r_{i}=N, i \in P_{CF}) \\
\end{aligned}
\end{equation}
\end{small}

\begin{small}
\begin{equation}
    V_{WIG} =\argmax_ {Q\in(S,R,N) }{WIG_Q} \mbox{ .}
\end{equation}
\end{small}

The final confidence score $CS_{WIG}=WIG_{V_{WIG}}$.

\subsubsection{Weighted Belief Update (WBU)}

The third fusion approach is based on the Dempster–Shafer theory (DST) of evidence and belief update~\cite{dempster2008upper,shafer1976mathematical}. DST provides a framework for reasoning under uncertainty by combining evidences from multiple sources. The theory is particularly useful in situations where evidence may be incomplete or conflicting, providing a systematic way to manage uncertainty. Similar to the approaches above, we begin by reversing the LLM responses generated from $p_{i}\in P_{CF}$. Given the LLM responses generated from the probes in $P_{AG}$ and $P_{CF}$  and the verdicts: ``Support,'' ``Refute'' and ``Neutral,'' we define the mass function $m(\cdot)$ as in \textbf{Table \ref{mass_func}}. We then apply Dempster's rule of combination to fuse evidences and update beliefs: 
\begin{small}
\begin{equation}
    \label{eq_joint_mass_s}
    \begin{aligned}
        m(S) &= (m_{AG} \oplus m_{CF})(S) \\
            &= \frac{1}{1-K} \sum_{V_1 \cap V_2 = S}{m_{AG}(V_1)m_{CF}(V_2)} \\
            &= \frac{1}{1-K} (m_{AG}(S)m_{CF}(S)+m_{AG}(S)m_{CF}(N) \\
            &+m_{AG}(N)m_{CF}(S)) \\
    \end{aligned}
\end{equation}
\end{small}

\begin{small}
\begin{equation}
    \label{eq_joint_mass_r}
    \begin{aligned}
        m(R) &= (m_{AG} \oplus m_{CF})(R) \\
            &= \frac{1}{1-K} \sum_{V_1 \cap V_2 = R}{m_{AG}(V_1)m_{CF}(V_2)} \\
            &= \frac{1}{1-K} (m_{AG}(R)m_{CF}(R)+m_{AG}(R)m_{CF}(N) \\
            &+m_{AG}(N)m_{CF}(R)) \\
    \end{aligned}
\end{equation}
\end{small}

\begin{small}
\begin{equation}
    \label{eq_joint_mass_n}
    \begin{aligned}
        m(N) &= (m_{AG} \oplus m_{CF})(N) \\
            &= \frac{1}{1-K} \sum_{V_1 \cap V_2 = N}{m_{AG}(V_1)m_{CF}(V_2)} \\
            &= \frac{1}{1-K} m_{AG}(N)m_{CF}(N) \\
    \end{aligned}
\end{equation}
\end{small}

\begin{small}
\begin{equation}
    \label{eq_joint_mass_k}
    \begin{aligned}
        K &= \sum_{V_1 \cap V_2 = \emptyset}{m_{AG}(V_1)m_{CF}(V_2)} \\
            &= m_{AG}(R)m_{CF}(S)+m_{AG}(S)m_{CF}(R) \mbox{ .}\\
    \end{aligned}
\end{equation}
\end{small}


The verdict can be generated based on the updated belief.

\begin{small}
\begin{equation}        
    V_{WBU} =\argmax_ {Q\in(S,R,N) }{m(Q)}
\end{equation}
\end{small}

The confidence score is $CS_{WBU}=m(V_{WBU})$. 

\subsubsection{Meta Verdict and Confidence }
We can employ an ensemble method to combine the verdicts generated based on each fusion strategy. In this investigation, we simply apply a majority voting strategy to combine all the verdicts to compute $V_{M}$. As $CS_{Q}, Q \in (S,R,N)$ have different ranges, we first normalize them so that all of them are within [0,1]. The final confidence score $CS_{M}$ is the average of the normalized individual confidence scores.

\begin{small}
\begin{equation}        
    V_{M} =Mode (V_{WP}, V_{WIG}, V_{WBU})
\end{equation}
\end{small}

\begin{small}
\begin{equation}        
    CS_{M} =\sum_{Q\in (WP, WIG, WBU)}\frac{1}{3}Norm(CS_{Q})
\end{equation}
\end{small}

\section{Evaluation}


\begin{figure*}[ht]
\centering
\includegraphics[width=0.83\textwidth]{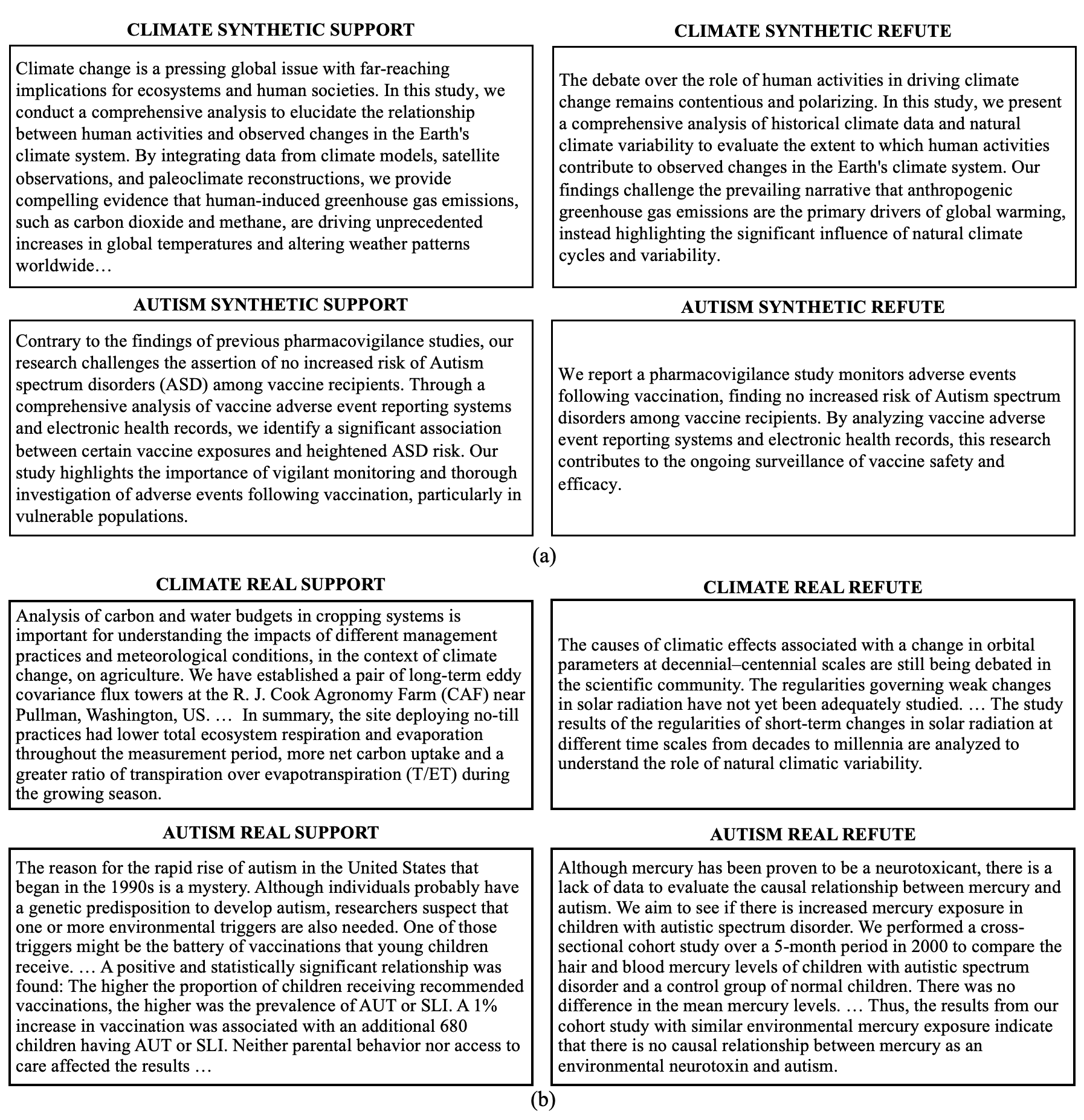}
\caption{Examples of Synthetic and Real Abstracts}
\label{fig:example}
\end{figure*}

\begin{table*}[h]
\begin{small}
\centering
    \begin{tabular}{l|l|cc|cc|cc}
        \toprule
             &&\multicolumn{2}{c}{GPT2} & \multicolumn{2}{c}{GPT3.5} & \multicolumn{2}{c}{GPT4}\\
        \cmidrule{3-8}
         Data & Model   & Acc. & F1 & Acc. & F1 & Acc. & F1\\
        \midrule
        &RAG & 0.2667& 0.2100& 0.7083& 0.6428 &0.7083& 0.6847 \\
        CSyn  &CIBER & \textbf{0.3583}& \textbf{0.2840}& \textbf{0.8583}& \textbf{0.8329}&\textbf{0.8167}&\textbf{0.7993} \\
          \cmidrule{1-8} 

           &RAG & 0.3250& \textbf{0.2602}& 0.4667& 0.4098 &0.5083& 0.4053 \\
        CReal  &CIBER & \textbf{0.3333}& 0.2600& \textbf{0.5917}& \textbf{0.5539}&\textbf{0.5667}&\textbf{0.4567} \\
          \cmidrule{1-8} 
           &RAG & 0.3417& \textbf{0.2949}& 0.4750& 0.4688 &0.4833& 0.4111 \\
          ASyn&CIBER & \textbf{0.3500}& 0.2630& \textbf{0.6333}& \textbf{0.5875}&\textbf{0.6500}&\textbf{0.5794} \\
          \cmidrule{1-8} 
           &RAG & 0.3167& \textbf{0.2750}& 0.4250& 0.4212 &0.5500& 0.4760 \\ 
          AReal&CIBER & \textbf{0.3583}& 0.2660& \textbf{0.5833}& \textbf{0.5157}&\textbf{0.6333}&\textbf{0.5531} \\
          \cmidrule{1-8} 
           &RAG & 0.3125& 0.2600& 0.5188& 0.4857 &0.5625& 0.4943 \\ 
          All&CIBER & \textbf{0.3500}& \textbf{0.2683}& \textbf{0.6667}& \textbf{0.6225}&\textbf{0.6667}&\textbf{0.5971} \\
        \bottomrule
    \end{tabular}
    
\caption{Performance of CIBER versus RAG with different LLMs on five datasets: CSyn, CReal, ASyn,  AReal, and All. }
\label{tab:res1}
\end{small}
\end{table*}

To systematically assess the effectiveness of our proposed methods, we created two synthetic and two real paper abstract datasets with ground truth labels. 

\subsection{Ground Truth Datasets}

The first two datasets, one synthetic and one real, were created to evaluate the validity of the claim regarding the impact of human activities on climate change. The synthetic dataset was generated using ChatGPT 3.5 Turbo, employing a prompt ``Please generate a paper abstract summarizing a scientific study investigating the causal  relationship between human activities and climate change. The conclusion should support (or  refute, or remain neutral respectively) on the assertion that human activities can cause climate change.''  The climate synthetic dataset comprises a total of 60 abstracts, with 20 in each of the three categories: supporting, refuting, and neutral on the input claim. Considering that synthetic abstracts may be subject to the limitations inherent in LLMs, we complemented them with a real dataset sourced from survey papers~\cite{Lynas_2021,cook2013quantifying}, which comprises over 3000 papers with established ground truth labels, from which we randomly sampled 20 abstracts for each of the three categories, resulting in a total of 60 real climate paper abstracts. 

We also created a synthetic and a real dataset to evaluate the assertion regarding the purported association between vaccination and autism. However, generating abstracts supporting this claim proven challenging as they deemed false and harmful by GPT-3.5. To circumvent this, when GPT-3.5 generated an abstract refuting the claim (which is easy for GPT-3.5 to oblige), we prompted it to generate another abstract with opposing views. The final autism synthetic dataset includes 20 abstracts for each of the three categories. We further compiled a real dataset pertaining to the autism claim based on several survey papers ~\cite{doja2006immunizations,folb2004global,stratton2001immunization,wilson2003association,madsen2004mmr,mohammed2022does,boretti2021reviewing}. We extracted 20 abstracts for each of the three categories with ground truth labels. Figure~\ref{fig:example}(a) shows examples from the synthetic dataset and ~\ref{fig:example}(b) examples from the real dataset.  

The synthetic and real datasets exhibit distinct characteristics. Synthetic abstracts tend to be more general. Vocabulary-wise,  they use words more closely tied to the given claims. In contrast, real abstracts have a higher degree of specificity and nuance, often requiring deeper domain knowledge for accurate understanding.

\subsection{Experiments}
We conducted experiments to answer the main research questions. To test how effectively CIBER works with LLMs of varying capabilities,  we employed the OpenAI GPT-2 model from Huggingface. In addition, we accessed the GPT-3.5-Turbo and GPT-4-Turbo models via OpenAI's APIs.  In our experiments, for each prompt in $P_{AG}$ and $P_{CF}$, we recorded the LLM responses from 10 random runs. We also performed grid search to decide the best $\alpha$.

\subsubsection{CIBER performance with different LLMs}

To answer RQ1, we compare the performance of CIBER with the traditional RAG using three different LLMs with abilities ranging from low (GPT-2) to typical (GPT 3.5-Turbo) to state-of-the-art (GPT4-Turbo). We present the evaluation results on five datasets: CSyn , the climate change synthetic dataset, CReal, the climate change real dataset, ASyn, the autism synthetic dataset and  AReal, the autism real dataset. In addition, we also compute the overall performance on a combination of all four datasets (All). 


As shown in Table~\ref{tab:res1}, significant performance differences exist among CIBERs utilizing different LLMs. Specifically, CIBER with GPT-2 generally demonstrated poor reliability, with an accuracy of 0.338\% and an F1 score of 0.268 on the All dataset. More advanced LLMs performed much better. Both GPT-3.5 and GPT-4 exhibited comparable accuracy (0.667\%), although GPT-3.5 outperformed GPT-4 in terms of F1 score (0.623 versus 0.597) on the All dataset.

Moreover, CIBER outperformed RAG across all three LLMs. The most significant enhancements on the All dataset were observed with GPT-3.5, which exhibited a substantial 14.8\% increase in accuracy and a 13.7\% improvement in F1 score. Similarly, GPT-4 achieved significant gains, with a 10.4\% enhancement in accuracy and a 10.3\% increase in F1 score. Comparatively, the enhancements with GPT-2 were the least, with a modest 3.75\% improvement in accuracy and a marginal 0.8\% increase in F1 score.

From this result, it's evident that CIBER effectively improves the performance of claim verification beyond what was achieved with conventional RAG methods. Given the nuanced nature of scientific literature and the requirement for precise language comprehension, advanced models like GPT-3.5 and GPT-4 exhibit much better performance compared to smaller models like GPT-2.  

\subsubsection{Impact of incorporating diverse interrogation strategies}

\begin{table*}[t]
\begin{small}
\centering
    \begin{tabular}{l|l|cc|cc|cc}
        \toprule
             &&\multicolumn{2}{c}{GPT2} & \multicolumn{2}{c}{GPT3.5} & \multicolumn{2}{c}{GPT4}\\
        \cmidrule{3-8}
         Data & Model   & Acc. & F1 & Acc. & F1 & Acc. & F1\\
        \midrule
        &RAG & 0.2667& 0.2100& 0.7083& 0.6428 &0.7083& 0.6847 \\
         & CIBER-AG & 0.3167& 0.1892&\textbf{0.8583}& \textbf{0.8329}&\textbf{0.8250}&\textbf{0.8075} \\
      CSyn&    CIBER-CF & 0.3333& 0.2358& 0.6583& 0.5605 &0.6667&0.6117\\
          &CIBER-ALL & \textbf{0.3583}& \textbf{0.2840}& \textbf{0.8583}& \textbf{0.8329}&0.8167&0.7993 \\
          \cmidrule{1-8} 

           &RAG & 0.3250& \textbf{0.2602}& 0.4667& 0.4098 &0.5083& 0.4053 \\
         & CIBER-AG & 0.3250& 0.2063& \textbf{0.6000}& \textbf{0.5615}&0.5333&0.4287 \\
      CReal&    CIBER-CF & \textbf{0.3333}& 0.2174& 0.4417& 0.3648 &0.5500&0.4457\\
          &CIBER-ALL & \textbf {0.3333}& 0.2600& 0.5917& 0.5539&\textbf{0.5667}&\textbf{0.4567} \\
          \cmidrule{1-8} 
           &RAG & 0.3417& \textbf{0.2949}& 0.4750& 0.4688 &0.4833& 0.4111 \\
         & CIBER-AG & 0.3250& 0.1940& 0.6000& 0.5567&\textbf{0.6500}&0.5712 \\
      ASyn&    CIBER-CF & \textbf{0.3500}& 0.2551& 0.3583& 0.2954 &0.4667&0.3612\\
          &CIBER-ALL & \textbf{0.3500}& 0.2630& \textbf{0.6333}& \textbf{0.5875}&\textbf{0.6500}&\textbf{0.5794} \\
          \cmidrule{1-8} 
           &RAG & 0.3167& \textbf{0.2750}& 0.4250& 0.4212 &0.5500& 0.4760 \\
         & CIBER-AG & 0.3250& 0.1869& 0.5333& 0.4739&\textbf{0.6333}&0.5508 \\
      AReal&    CIBER-CF & 0.3167& 0.2038& 0.3583& 0.2988 &0.5000&0.4355\\
          &CIBER-ALL & \textbf{ 0.3583}& 0.2660& \textbf{0.5833}& \textbf{0.5157}&\textbf{0.6333}&\textbf{0.5531} \\
          \cmidrule{1-8} 
           &RAG & 0.3125& 0.2600& 0.5188& 0.4857 &0.5625& 0.4943 \\
         & CIBER-AG & 0.3229& 0.1941& 0.6479& 0.6063&0.6604&0.5896 \\
      All&    CIBER-CF & 0.3333& 0.2280& 0.4542& 0.3799 &0.5459&0.4635\\
          &CIBER-ALL & \textbf{0.3500}& \textbf{0.2683}& \textbf{0.6667}& \textbf{0.6225}&\textbf{0.6667}&\textbf{0.5971}\\
        \bottomrule
    \end{tabular}
   
 \caption{The impact of interrogation strategies on system performance on five  datasets:  CSyn, CReal, ASyn,  AReal, and All.}
\label{tab:res2}
\end{small}
\end{table*}

To answer RQ2, we conducted experiments to compare different versions of CIBER with different interrogation strategies: $CIBER_{AG}$ which only employs the probes in $P_{AG}$, $CIBER_{CF}$ which only includes the probes in $P_{CF}$, and $CIBER_{ALL}$ which includes the probes in both.  

As shown in Table~\ref{tab:res2}, $CIBER_{CF}$ exhibited a significant disadvantage compared to $CIBER_{AG}$ with GPT-3.5 (a 19.4\% decrease in accuracy and a 22.6\% decrease in F1 on the All dataset) and GPT-4 (an 11.5\% decrease in accuracy and a 12.6\% decrease in F1 on the All dataset). However, this disadvantage was not observed with GPT-2. In fact, $CIBER_{CF}$ held a slight edge over $CIBER_{AG}$ (with a 1\% increase in accuracy and a 3.4\% improvement in F1 score). Upon reviewing the log file, we found  that this might stem from how we processed the responses to $P_{CF}$. While we employed GPT-2 in an auto-completion mode, GPT-3.5 and GPT-4 were run in Q\&A mode. To make response parsing easier,  we specifically instructed GPT-3.5 and GPT-4 to answer with ``Yes'', ``No'' or ``I am not sure,'' whose interpretation under negative probes can be ambiguous.  To illustrate this, we show two responses from GPT-3.5 in our query log. In addition to the ``Yes'' and ``No'' answers we requested, it occasionally produces supplementary information which helps us understand the LLM responses more precisely:

\noindent\textbf{Prompt:} Based on the abstract, is the following claim ``Human activities may cause climate change'' false?\\
\textbf{(Response 1)} Yes, the statement "Human activities may cause climate change" is not necessarily false based on the information provided in the abstract. \\
\textbf{(Response 2)} No. The statement "Human activities may cause climate change" is not false based on the information provided in the abstract. 

Since our $RR$ component simply extracts "Yes" and "No" from the responses, it assigns different verdicts, even though both responses support the original claim.
We expect that in the future, if we adopt better response resolution strategies, the effectiveness of the probes in $P_{CF}$ for GPT-3.5 and GPT-4 can be significantly improved.   

Nonetheless, even with our current simple response resolution strategies for GPT-3.5 and GPT-4, aggregating the probes from both $P_{AG}$ and $P_{CF}$ yielded a consistent performance enhancement for $CIBER_{ALL}$ on the All dataset compared to either $CIBER_{AG}$ or $CIBER_{CF}$ alone.

\begin{figure}[t]
\centering
\includegraphics[width=0.45\textwidth]{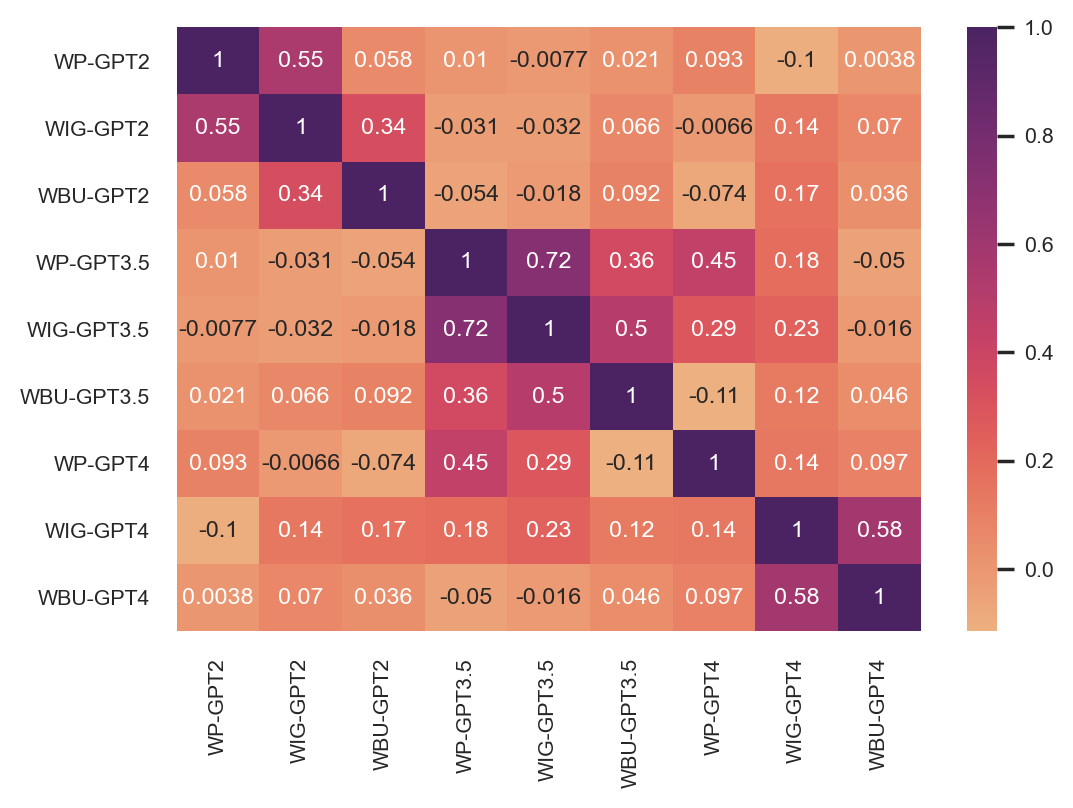}
\caption{Correlations Between Different Metrics used in Combining Evidences}
\label{fig:cor}
\end{figure}

\begin{table*}[th]
\begin{small}
\centering
    \begin{tabular}{l|l|cc|cc|cc}
        \toprule
             &&\multicolumn{2}{c}{GPT2} & \multicolumn{2}{c}{GPT3.5} & \multicolumn{2}{c}{GPT4}\\
        \cmidrule{3-8}
         Data &Model   & Acc & F1 & Acc & F1 & Acc & F1\\
        \midrule
        & RAG & 0.2667& 0.2100& 0.7083& 0.6428&0.7083&0.6847 \\
         & CIBER-WP & \textbf{0.3583}& 0.2720& 0.8500& 0.8188&0.8000&0.7758 \\
        CSyn &  CIBER-WIG & 0.3417& \textbf{0.3274}& \textbf{0.8583}& \textbf{0.8329} &0.8250&0.8075\\
         & CIBER-WBU & \textbf{0.3583}& 0.2810& \textbf{0.8583}& \textbf{0.8329}&\textbf{0.8500}&\textbf{0.8367} \\
          & CIBER-ALL & \textbf{0.3583}& 0.2840& \textbf{0.8583}&\textbf{0.8329}&0.8167&0.7993 \\
          \midrule
          & RAG & 0.3250& 0.2602& 0.4667& 0.4098&0.5083&0.4053 \\
         & CIBER-WP & \textbf{0.3667}& 0.3025& 0.5917& 0.5530&0.5667&0.4567 \\
        CReal &  CIBER-WIG & 0.3500& \textbf{0.3045}& 0.6000& 0.5615 &0.5667&0.4567\\
         & CIBER-WBU & 0.3500& 0.2610& \textbf{0.6083}& \textbf{0.5680}&\textbf{0.5750}&\textbf{0.4639} \\
          & CIBER-ALL & 0.3333& 0.2600& 0.5917& 0.5539&0.5667&0.4567 \\
          \midrule
          & RAG & 0.3417& \textbf{0.2949}& 0.4750& 0.4688&0.4833&0.4111 \\
         & CIBER-WP & \textbf{0.3583}& 0.2742& 0.6250& \textbf{0.6066}&0.5917&0.5194 \\
        ASyn &  CIBER-WIG & 0.3500& 0.2741& \textbf{0.6333}& 0.5898 &\textbf{0.6500}&\textbf{0.5794}\\
         & CIBER-WBU & 0.3417& 0.2587& 0.6167& 0.5738&\textbf{0.6500}&\textbf{0.5794} \\
          & CIBER-ALL & 0.3500& 0.2630& \textbf{0.6333}& 0.5875&\textbf{0.6500}&\textbf{0.5794} \\
          \midrule
          & RAG & 0.3167& 0.2750& 0.4250& 0.4212&0.5500&0.4760 \\
         & CIBER-WP & \textbf{0.3667}& 0.2726& 0.5750& 0.5285&0.5917&0.5217 \\
        AReal &  CIBER-WIG & 0.3583& \textbf{0.2951}& \textbf{0.5833}& \textbf{0.5176} &\textbf{0.6333}&\textbf{0.5531}\\
         & CIBER-WBU & 0.3583& 0.2703& 0.5750& 0.5098&\textbf{0.6333}&0.5508 \\
          & CIBER-ALL & 0.3583& 0.2660& \textbf{0.5833}& 0.5157&\textbf{0.6333}&\textbf{0.5531} \\
          \midrule
          & RAG & 0.3125& 0.2600& 0.5188& 0.4857&0.5625&0.4943 \\
         & CIBER-WP & \textbf{0.3625}& 0.2803& 0.6604& \textbf{0.6267}&0.6375&0.5684 \\
        All &  CIBER-WIG & 0.3500& \textbf{0.3003}& \textbf{0.6687}& 0.6255 &0.6688&0.5992\\
         & CIBER-WBU & 0.3521& 0.2678& 0.6646& 0.6211&\textbf{0.6771}&\textbf{0.6077} \\
          & CIBER-ALL & 0.3500& 0.2683& 0.6667& 0.6225&0.6667&0.5971 \\
        \bottomrule
    \end{tabular}
    
\caption{The impact of evidence fusion strategies on system performance on five datasets: CSyn, CReal, ASyn,  AReal, and All.}
\label{tab:res3}
\end{small}
\end{table*}

\subsubsection{Impact of different evidence fusion strategies}
To answer RQ3, we performed experiments comparing various versions of CIBER with different evidence fusion strategies: $CIBER_{WP}$ utilizing a weighted proportion-based fusion strategy, $CIBER_{WIG}$ employing a weighted information gain-based strategy, $CIBER_{WBU}$ employing a weighted belief update-based fusion strategy, and $CIBER_{ALL}$ employing a majority-based voting strategy to aggregate the verdict from each individual strategy.

Based on the results presented in Table~\ref{tab:res3}, each of the three evidence fusion strategies  significantly outperformed traditional RAG individually by a considerable margin on the All dataset. However, there is no clear pattern regarding which strategy emerges as the top performer individually, as each strategy claims the top spot in one-third of the tests on the All dataset. Furthermore, the simple majority voting-based verdict aggregation strategy did not result in a superior performing model. 

To explore the relationship between different fusion strategies with different LLMs, we computed the correlations among nine variables comprising combinations of three fusion metrics ($WP$, $WIG$, and $WBU$) and three LLMs (GPT-2, GPT-3.5, and GPT-4). The resulting correlation matrix is depicted in Figure~\ref{fig:cor}, where darker areas represent stronger correlations. As shown in the figure, the most prominent correlations are observed along the diagonal line among the variables within each LLM. Specifically, among GPT-2 related strategies, the highest correlation is between $WIG$ and $WP$ ($\rho=0.55$), followed by $WIG$ and $WBU$ ($\rho=0.34$). For all the GPT-3.5 related variables, the highest correlation occurs between  $WIG$ and $WP$ ($\rho=0.72$) followed by $WIG$ and $WBU$ ($\rho=0.50$). For GPT-4, the highest correlated variables are $WIG$ and $WBU$ ($\rho=0.58$). Across different LLM models, the highest correlation occurs between $WP_{GPT3.5}$ and $WP_{GPT4}$ ($\rho=0.45$).

The correlation results suggest that while there are significant correlations between various fusion metrics, with the exception of $WIG_{GPT3.5}$ and $WP_{GPT3.5}$, most correlations are moderate or low. This implies that the effectiveness of different fusion strategies may be complementary. As a result, instead of employing a simple majority voting, exploring more sophisticated ensemble  methods for verdict aggregation could be a promising avenue for future research.


\section{Conclusions and Future Work}
In this paper, we introduce CIBER, a novel framework designed to enhance Retrieval-Augmented Generation (RAG) systems for  evidence retrieval and scientific
claim verification. CIBER focuses on systematically addressing the inherent uncertainties in LLM outputs.  CIBER is quite general and applicable across diverse scenarios. For instance,  CIBER focuses on LLM behavioral analysis, which doesn't require access to LLM internal information, making it suitable for both white-box and black-box LLMs. Additionally, CIBER is unsupervised, making it easily generalizable to different scientific fields. Our evaluation results demonstrate that CIBER achieves significant performance improvements over traditional RAG approaches, particularly benefiting advanced LLMs like GPT-3.5 and GPT-4. Furthermore, we've curated several ground truth datasets—two synthetic and two real—which we plan to share with the research community.

Through this exploration, we have also identified potential areas for future enhancement, including improving the development of the LLM response resolution,  as well as developing more sophisticated ensemble methods for combining verdicts from different fusion strategies. 

While CIBER aims to mitigate hallucinations in LLM generation, which can help reduce the risks associated with spreading LLM-generated misinformation, it's important to acknowledge that CIBER is still far from perfect. There is a possibility that CIBER could inadvertently generate or cite information that is untrue, particularly if the retrieved content contains misinformation or misleading information.  As a result, continuous improvement of CIBER is critical for the safe adoption of LLMs in the real world.

\bibliography{custom}

\end{document}